# An Unsupervised Dynamic Image Segmentation using Fuzzy Hopfield Neural Network based Genetic Algorithm

Amiya Halder[1] and Soumajit Pramanik[2]

[1] Computer Science & Engineering, St. Thomas' College of Engineering & Technology
Kolkata, West Bengal, India
*amiya_halder@indiatimes.com*

[2] Computer Science & Engineering, St. Thomas' College of Engineering & Technology
Kolkata, West Bengal, India
*Soumajit.pramanik@gmail.com*

**Abstract**
This paper proposes a Genetic Algorithm based segmentation method that can automatically segment gray-scale images. The proposed method mainly consists of spatial unsupervised gray-scale image segmentation that divides an image into regions. The aim of this algorithm is to produce precise segmentation of images using intensity information along with neighborhood relationships. In this paper, Fuzzy Hopfield Neural Network (FHNN) clustering helps in generating the population of Genetic algorithm which there by automatically segments the image. This technique is a powerful method for image segmentation and works for both single and multiple-feature data with spatial information. Validity index has been utilized for introducing a robust technique for finding the optimum number of components in an image. Experimental results shown that the algorithm generates good quality segmented image.

***Keywords:*** *Clustering, Image Segmentation, Fuzzy Hopfield Neural Network, Genetic Algorithm.*

## 1. Introduction

Segmentation refers to the process of partitioning a digital image into multiple segments or regions. The goal of segmentation is to simplify the representation of an image into something that is more meaningful and easier to analyze. Image segmentation is typically used to locate objects and boundaries in images [1]. More precisely, image segmentation is the process of assigning a label to every pixel in an image such that pixels with the same label share certain visual characteristics. Image segmentation is a very important field in image analysis object recognition, image coding and medical imaging. Segmentation is very challenging because of the multiplicity of objects in an image and the large variation between them. Image segmentation is the process of division of the image into regions with similar attributes. In many object based image segmentation applications, the number of cluster is known a priori, but our proposed scheme is automatically determined the number of cluster which is produced the segmentation of images. The proposed technique should be able to provide good results whereas K-means algorithm which may get stuck at values which are not optimal [2]. Some of the several unsupervised clustering algorithms developed include K-means [3,4], ISODATA [5], self-organizing feature map (SOM) [6], Particle Swarm Optimization (PSO) [7], learning vector quantizers (LVQ) [8], GA based Clustering [9,10] etc.

One natural view of segmentation is that we are attempting to determine which components of a data set naturally "belong together". Clustering is a process whereby a data set is replaced by clusters, which are collections of data points that "belong together". Thus, it is natural to think of image segmentation as image clustering i.e. the representation of an image in terms of clusters of pixels that "belong together". The specific criterion to be used depends on the application. Pixels may belong together because of the same color or similarity measure. The result of this algorithm produced a better result to compare with other techniques. Various segmentation techniques have been developed for image segmentation [11-14].

This paper presents in fact a possible extension of the previously presented concept of GA based segmentation [13, 14] which was based on the GA based clustering of gray level images. In previous technique, the chromosome of GA is selected by randomly chosen in the image of the pixel. In this paper, fuzzy Hopfield neural network helps in generating the population of genetic algorithm which there by automatically segments the image [15]. This paper describes the concepts of optimal clustering technique which gives the better results compare to other state-of-the-art image segmentation methods and existing segmentation techniques [11,13,14,15,19].

The rest of this paper is organized as follows: - Section 2, the concepts of clustering is provided. Section 3 gives the concepts of Genetic algorithm. Section 4 gives overview of FHNN based Genetic Algorithm clustering approach. Section 5 describes the proposed algorithm and section 6 described the experimental results and section 7 concludes the paper.

## 2. Clustering

The process of grouping a set of physical or abstract objects into classes of similar objects is called clustering. A cluster is a collection of data objects that are similar to one another within the same cluster and are dissimilar to the objects in other clusters. By clustering, one can identify dense and sparse regions and therefore, discover overall distribution patterns and interesting correlations among data attributes.

Clustering may be found under different names in different contexts, such as unsupervised learning (in pattern recognition), numerical taxonomy (in biology, ecology), typology (in social sciences) and partition (in graph theory) [3]. By definition, "cluster analysis is the art of finding groups in data", or "clustering is the classification of similar objects into different groups, or more precisely, the partitioning of a data into subsets (clusters), so that the data in each subset (ideally) share some common trait- often proximity according to some defined distance measure" [4]. Clustering is a challenging field of research as it can be used as a stand-alone tool to gain insight into the distribution of data, to observe the characteristics of each cluster, and to focus on a particular set of clusters for further analysis. Alternatively, cluster analysis serves as a pre-processing step for other algorithms, such as classification, which would then operate on detected clusters.

Hierarchical agglomerative clustering techniques start with as many clusters as there are unique values. Then pairs of cluster are successively merged till the optimal number of clusters is reached, depending on the termination condition. Termination condition is to be chosen carefully; else the hierarchical agglomerative clustering technique will ultimately yield one cluster containing all the values [17].
Clustering is a useful unsupervised data mining technique which partitions the input space into K regions depending on some similarity/dissimilarity metric where the value of K may or may not be known a priori. The main objective of any clustering technique is to produce a K×n partition matrix U(X) of the given data set X, consisting of n patterns, X =(x1, x2. . . xn )[18].

## 3. Genetic Algorithm

Genetic Algorithm (GA) is a population-based stochastic search procedure to find exact or approximate solutions to optimization and search problems. Modeled on the mechanisms of evolution and natural genetics, genetic algorithms provide an alternative to traditional optimization techniques by using directed random searches to locate optimal solutions in multimodal landscapes [16]. Each chromosome in the population is a potential solution to the problem. Genetic Algorithm creates a sequence of populations for each successive generation by using a selection mechanism and uses operators such as crossover and mutation as principal search mechanisms - the aim of the algorithm being to optimize a given objective or fitness function. An encoding mechanism maps each potential solution to a chromosome. An objective function or fitness function is used to evaluate the ability of each chromosome to provide a satisfactory solution to the problem. The selection procedure, modeled on nature's survival-of-the-fittest mechanism, ensure that the fitter chromosomes have a greater number of offspring in the subsequent generations. For crossover, two chromosomes are randomly chosen from the population. Assuming the length of the chromosome to be $l$, this process randomly chooses a point between 1 and $l$-1 and swaps the content of the two chromosomes beyond the crossover point to obtain the offspring. A crossover between a pair of chromosomes is affected only if they satisfy the crossover probability.

Mutation is the second operator, after crossover, which is used for randomizing the search. Mutation involves altering the content of the chromosomes at a randomly selected position in the chromosome, after determining if the chromosome satisfies the mutation probability.

In order to terminate the execution of GA, a stopping criterion is specified. Specifying the number of iterations of the generational cycle is one common technique of achieving this end.

## 4. FHNN based Genetic Algorithm Clustering

The searching capability of GAs can be used for the purpose of appropriately clustering a set of n unlabeled points in N-dimension into K clusters [9]. In this proposed scheme, the same idea can be applied on image data. We consider a gray level image of size m×n. The basic steps of the GA-clustering algorithm for clustering image data are as follows:

## 4.1 Encoding

Each chromosome represents a solution which is a sequence of K cluster centers. For an N-dimensional space, each cluster center is mapped to N consecutive genes in the chromosome. For image datasets each gene is an integer representing an intensity value.

## 4.2 Population initialization

Image segmentation can be considered as a clustering process in which the pixels are classified to the specific regions based on their gray-level values and spatial connectivity. In addition to the neural network-based technique, the fuzzy set has also been demonstrated to address segmentation problems. A new segmentation method using a fuzzy Hopfield neural network (FHNN) based upon the pixel classification for image segmentation is described here. This approach is different from the previous ones, in that a fuzzy reasoning strategy is added into a neural network. In FHNN, the problem of the image segmentation is regarded as a process of the minimization of a cost function. This cost function is defined as the Euclidean distance between the gray levels in a histogram to the cluster centers represented in the gray levels. The structure of this network is constructed as a 2-D fully interconnected array with the columns representing the number of classes and the rows representing the gray level of pixels taken as training samples [20]. However, a training sample does not necessarily belong to one class. Instead, a certain degree of class membership is associated with every sample. In FHNN, an original Hopfield network is modified and the fuzzy c-means clustering strategy is added. Consequently, the energy function may be quickly converged into a local minimum, in order to produce a satisfactory resulting image. Compared with conventional techniques, the major strength of the presented FHNN is that it is computationally more efficient due to the inherent parallel structures.

The network consists of n×c neurons which are fully interconnected neurons. Let $V_{x,i}$ denote the binary state of neuron (x,i) and $W_{x,i;y,j}$ be the interconnection weight between the neuron (x, i) and the neuron (y, j). A neuron (x, i) receives each neuron (y, j) with $W_{x,i;y,j}$ and a bias $I_{x,i}$ from outside, which can then be expressed by

$$Net_{x,i} = \sum_{y=1}^{n}\sum_{j=1}^{c} W_{x,i;y,j} + I_{x,i} \quad (1)$$

and the Lyapunov energy function of the two-dimensional Hopfield network is given by

$$E = \frac{1}{2}\sum_{x=1}^{n}\sum_{y=1}^{n}\sum_{i=1}^{c}\sum_{j=1}^{c} V_{x,i} W_{x,i;y,j} V_{y,j} - \sum_{x=1}^{n}\sum_{i=1}^{c} I_{x,i} V_{x,i} \quad (2)$$

Each column of the Hopfield neural network represents a class and each row represents samples in a proper class. The network reaches a stable state when the Lyapunov energy function is minimized. For example, a neuron (x,i) in a firing state (i.e., $V_{x,i}=1$) indicates that sample $z_x$, belongs to class i. But, in the fuzzy Hopfield neural network, a neuron (z,i) in a fuzzy state indicates that sample $z_x$, belongs to class i with a degree of uncertainty described by a membership function.

Each pixel in an L×L image can be considered as an object being assigned to one of M labels. Then, the constraint satisfaction neural network consists of L×L×M neurons that can be conceived as a 3-D array for the image-segmentation problem. The number of neurons is dependent on image size; the larger the image size, the more neurons that are required. Here the global histogram, but not the spatial connectivity information of the medical images, is employed for the process of the image segmentation.

In a 2-D image, each pixel is assigned one of n gray levels. If the number of sub-regions c is defined in advance, then the FHNN consists of n×c neurons that can be conceived as a 2-D array. Consequently, the number of neurons is independent of the image size. In this section, we will show that the medical image segmentation problem can be mapped onto a Hopfield neural network so that the cost function serves as the energy function of the network. The idea is to form the energy function of the network in terms of the intra-class energy function. In the pattern recognition application, the degree of natural association is expected to be high among members belonging to the same class and low among members of different clusters. In other words, the intra set (within-class) distance should be small. The proposed technique first assigns samples to their associated classes in such a manner that the Euclidean distance between arbitrary samples to their class center is minimized. This is referred to as the intra-class assignment. In linear discriminate analysis the concept of within-class scatter matrix is widely used for class separability. The iteratively updated synaptic weight between the neuronal interconnections will gradually force the network to converge into a stable state where its energy Function is minimized.

If we suppose that the number of classes, c ≥ 2, is pre-specified, and let Z = ($z_1$, $z_2$, $z_3$...$z_m$) be a set of samples to be classified. $M_{n×c}$ is the set of all real n×c matrices where a matrix $U = \{\mu_{x,i} \in M_{n \times c}\}$ is called a fuzzy-c partition if it satisfies the following conditions:

$$\mu_{x,i} \in [0,1], \text{ for all } x \text{ and } i$$

$$\sum_{i=1}^{c} \mu_{x,i} = 1, \text{ for all } x$$

$$0 < \sum_{x=1}^{n} \mu_{x,i} < n, \text{ for all } i$$

$$\sum_{x=1}^{n} \sum_{i=1}^{c} \mu_{x,i} = n$$

We will let the brightness of a given image be represented by n gray levels. The frequency of occurrence of each gray level $z_x$, will be denoted by $p_x$. Using the within-class scatter matrix criteria, the optimization problem can be mapped into a 2-D fully interconnected Hopfield neural network with the fuzzy c-means strategy for medical image segmentation. The total weighed input for neuron (x,i) and Lyapunov energy can be modified as

$$Net_{x,i} = \left[ z_x - \sum_{y=1}^{n} W_{x,i;y,j}(\mu_{y,i})^m \right]^2 + I_{x,i} \quad (3)$$

$$E = \frac{1}{2} \sum_{x=1}^{n} \sum_{i=1}^{c} p_x(\mu_{y,i})^m \left[ z_x - \sum_{y=1}^{n} W_{x,i;y,j}(\mu_{y,j})^m \right]^2 -$$

$$\sum_{x=1}^{n} \sum_{i=1}^{c} I_{x,i}(\mu_{y,i})^m \quad (4)$$

is the total weighed input received from the neuron (y, i) in row i, $\mu_{x,i}$ is the output state at neuron (x,i), and m is the fuzzification parameter. Each column of this modified Hopfield network represents a class and each row represents a sample (gray level) in a proper class. The network reaches a stable state when the modified Lyapunov energy function is minimized.

**FHNN Algorithm:**

Step 1: Input a set of training gray levels Z = { $z_1$, $z_2$, . . . , $z_n$} and their associated frequencies of occurrence P = { $p_1$, $p_2$,.., $p_n$}, fuzzification parameter m (1≤ m < ∞), the number of classes c, and randomly initialize the states for all neurons U = [$\mu_{y,i}$] (membership matrix).
Step 2: Compute the weighted matrix W = [$W_{x,i;y,j}$] using

$$W_{x,i;y,j} = \frac{1}{\sum_{h=1}^{n} p_h(\mu_{h,i})^m p_y z_y} \quad (5)$$

Step 3: Calculate the input to each neuron (x, i)

$$Net_{x,i} = \left[ z_x - \sum_{y=1}^{n} \frac{1}{\sum_{h=1}^{n} p_h(\mu_{h,i})^m} p_y z_y (\mu_{y,i})^m \right]^2 \quad (6)$$

as $I_{x,i}=0$;

Step 4: Apply $\mu_{x,i} = \left[ \sum_{j=1}^{c} \left( \frac{Net_{x,i}}{Net_{x,j}} \right)^{\frac{2}{m-1}} \right]^{-1}$ for all i, to update the neuron's membership value in a synchronized manner;

Step 5: Compute Δ = max [|U(t+1)-Ut|]. if Δ> ε ,then go to Step 2, otherwise stop the process.
In Step 3, the inputs are calculated for all neurons. In Step 4, the fuzzy c-means clustering method is applied to determine the fuzzy state with the synchronous process. Here, a synchronous iteration is defined as an updated fuzzy state for all neurons using a software simulation.

4.3 Fitness computation

The fitness computation is accomplished in two steps. First, the pixel dataset is clustered according to the centers encoded in the chromosome under consideration, such that each intensity value $x_i$, i = 1, 2, ...,m× n is assigned to cluster with center $z_j$, j = 1, 2, ..., K.

$$\text{if } \| x_i - z_j \| < \| x_i - z_p \|, \quad (7)$$

$$\text{where } p = 1,2,..K, \text{ and } p \neq j,$$

The next step involves adjusting the values of the cluster centers encoded in the chromosome, replacing them by the mean points of the respective clusters. The new center $z_i^*$ for the cluster $C_i$ is gives by

$$z_i^* = \frac{1}{n_i} \sum_{x_j \in C_j} x_j, i = 1,2,..K \quad (8)$$

Subsequently, the clustering metric M is computed as the sum of Euclidean distances of each point from their respective cluster centers given by

$$M = \sum_{i=1}^{K} M_i \quad (9)$$

$$M_i = \sum_{x_j \in C_i} \| x_j - z_i \| \quad (10)$$

The fitness function is defined as

f = 1/M  (11)

A low value of intra-cluster spread is a characteristic of efficient clustering. Hence our objective is to minimize the clustering metric M i.e. maximize f.

4.4 Selection

This fitness level is used to associate a probability of selection with each individual chromosome. We apply Roulette Wheel selection, a proportional selection algorithm where the no. of copies of a chromosome that go into the mating pool for subsequent operations, is proportional to its fitness. If $f_i$ is the fitness of individual $C_i$ in the population, its probability of being selected is,

$$p_i = \frac{f_i}{\sum_{j=1}^{N} f_j}$$

Where N is the number of individuals in the population.

4.5 Crossover

In this paper, a single-point crossover with a fixed crossover probability of $\mu_c$ is used. The procedure followed is the same as that described in section 3.

4.6 Mutation

Each chromosome undergoes mutation with a fixed probability $\mu_m$. A number δ in the range [0, 1] is generated with uniform distribution. If the value at a gene position is v, after mutation it becomes
  v ± δ*v, v≠0
  v ± δ,   v=0

4.7 Termination criterion

We execute the processes of fitness computation, selection, crossover, and mutation for a predetermined number of iterations. In every generational cycle, the fittest chromosome till the last generation is preserved - elitism. Thus on termination, this chromosome gives us the best solution encountered during the search.

The algorithm is a two pass process in each iteration. In the first pass standard FHNN algorithm is used to generate the population. In the second pass, Genetic algorithm is applied on the population generated by the FHNN algorithm. Then cluster validity index for the fittest chromosome for particular value of K is computed using equation (12), in order to determine the validity of the clustering on the given dataset and the fittest chromosome for that K value is preserved. GA is performed for K=2 to $K_{max}$.

## 5. Validity Index

The cluster validity measure used in the paper is the one proposed by Turi [11]. It aims at minimizing the validity index given by the function,

$$V = y \times \frac{intra}{inter} \quad (12)$$

The term intra is the average of all the distances between each pixel x and its cluster centroid zi which is defined as

$$intra = \frac{1}{N} \sum_{i=1}^{K} \sum_{x \in C_i} \| x - z_i \|^2 \quad (13)$$

This term is used to measure the compactness of the clusters. The inter term is the minimum distances between the cluster centroids which is defined as

$$inter = \min(\| z_i - z_j \|^2), \quad (14)$$

$$where\ i = 1,2,...K-1,\ j = i+1,..,K$$

This term is used to measure the separation of the clusters. Lastly, y is given as

$$y = c \times N(2,1) + 1 \quad (15)$$

Where c is a user specified parameter and N(2,1) is a Gaussian distribution function with mean 2 and standard deviation 1, where the variable is the cluster number and is given as

$$N(\mu,\sigma) = \frac{1}{\sqrt{2\Pi\sigma^2}} e^{\frac{(k-\mu)^2}{2\sigma^2}}$$

Where k is the cluster number and μ=2 and σ=1 as per Turi's thesis on clustering [11]. This validity measure serves the dual purpose of
- minimizing the intra-cluster spread, and
- maximizing the inter-cluster distance.

Moreover it overcomes the tendency to select a smaller number of clusters (2 or 3) as optimal, which is an inherent limitation of other validity measures such as the Davies-Bouldin index or Dunne's index.

## 6. Experimental Results

The proposed algorithm has been applied to well known natural images such as Lena, mandrill and peppers etc. All the results have been reported in Figure 1 and Figure 2. These results have been compared to those of FHNN [15], DCPSO [19] and Dynamic GA [13,14]. The assumptions used for the implementation of the proposed algorithm are given as follows. The value of the parameter, c, for the validity index referred to from [11], is set to 25. The size of the population, P, is taken as 30, crossover rate, $\mu_c$, as 0.9 and mutation rate, $\mu_m$, as 0.01 [16]. The algorithm uses number of iterations as the terminating condition and it is set to 20. The value of $K_{max}$ is empirically set for the individual images. The maximum intensity level for each picture element in the image is taken as 255.

We have been compared to the results of validity index with the state-of-the-art image segmentation methods to DCPSO, Dynamic GA and FHNN in Figure 1. We consider the no. of cluster (output) of the proposed method is applying as an input of above state-of-the-art image segmentation methods. The result shows that the same no. of cluster is used in FHNN, DCPSO and Dynamic GA, FHNN and proposed method but our proposed technique gives the better results.

Figure 2 is shown the some segmented output images using DCPSO, FHNN, Dynamic GA and Proposed Method.

## 7. Conclusions

This paper presented a new approach for unsupervised image segmentation algorithm based on clustering technique which determines the optimal clustering of an image dataset, with minimum user intervention. In this paper, that the user does not need to predict the optimal number of clusters, required to partition the dataset, in advance. Comparison of the experimental results with that of other clustering methods, show that the technique gives satisfactory results when applied on well known natural images. Moreover results of its use on images from other fields (MRI, Satellite Images) demonstrate its wide applicability.

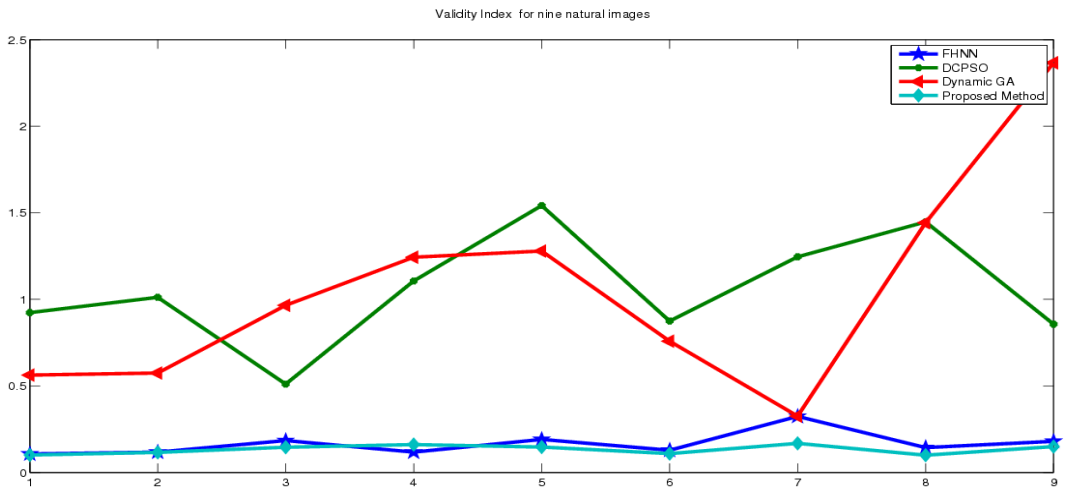

Figure 1: Validity index curve using FHNN, DCPSO, Dynamic GA based clustering and Proposed Method

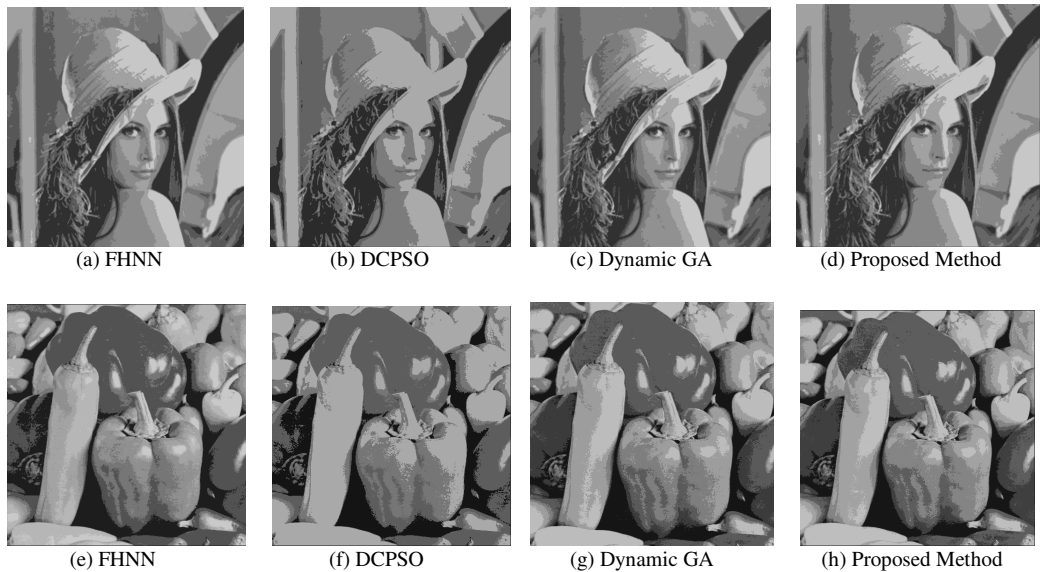

| (a) FHNN | (b) DCPSO | (c) Dynamic GA | (d) Proposed Method |
| (e) FHNN | (f) DCPSO | (g) Dynamic GA | (h) Proposed Method |

Figure 2: Segmented image using FHNN, DCPSO, Dynamic GA based clustering and Proposed Method

**Mr. Amiya Halder** received the B.Sc degree with Honours in Physics and B. Tech. in Computer Science & Engineering from the University of Calcutta, West Bengal, India, in 1998 and 2001, respectively. He obtained the M.E. degree in Computer Science & Engineering from the Jadavpur University, West Bengal, India, in 2003. At present, he is doing his Ph.D. in the field of Video Processing from Jadavpur University, Kolkata, West Bengal. Mr. Halder is a Asst. Prof. of the Department in Computer Science & Engineering, St. Thomas' College of Engineering & Technology,

Kolkata, West Bengal, India. He has authored or co-authored over 18 journal and conference papers. His area of interest is in the subject of Image Processing, Soft Computing and Video Processing.

**Soumajit Pramanik** received the B.E degree in Computer Science & Engineering from the St. Thomas' Colllege of Engg. & Tech., West Bengal, India, in 2011. At present, he is studying M. Tech. in ISI, Kolkata, West Bengal. He has authored or co-authored over 3 journal and conference papers. His area of interest is in the subject of Image Processing, Soft Computing.